\pdfoutput=1

\documentclass[11pt]{article}

\usepackage[preprint]{acl}

\usepackage{times}
\usepackage{latexsym}
\usepackage{booktabs}
\usepackage{enumitem}

\usepackage[T2A,T1]{fontenc}

\usepackage[utf8]{inputenc}
\usepackage[russian,german,english]{babel}

\usepackage{microtype}

\usepackage{inconsolata}

\usepackage{graphicx}
\newcommand{\comment}[1]{}
\title{Explaining novel senses using definition generation \\with open language models}

\author{
 \textbf{Mariia Fedorova\textsuperscript{1}},
 \textbf{Andrey Kutuzov\textsuperscript{1}},
 \textbf{Francesco Periti\textsuperscript{2}},
 \textbf{Yves Scherrer\textsuperscript{1}}
\\
 \textsuperscript{1}University of Oslo, Norway,
 \textsuperscript{2}KU Leuven - Flanders Make, Belgium
\\
 \small{
   \textbf{Correspondence:} \href{mailto:mariiaf@ifi.uio.no}{mariiaf@ifi.uio.no}
 }
}

\begin{document}
\maketitle
\begin{abstract}

We apply definition generators based on open-weights Large Language Models (LLMs) to the task of explaining novel word senses, taking target word usages as an input. To this end, we employ the datasets from the AXOLOTL'24 shared task on explainable semantic change modeling, which features Finnish, Russian and German languages. We fine-tune and provide publicly open-source models performing higher than the best submissions of the aforementioned shared task, which employed closed proprietary LLMs. In addition, we find that encoder-decoder definition generators perform on par with their decoder-only counterparts.

\end{abstract}

\section{Introduction and related work}
\label{sec:intro}

Recent NLP advancements have sparked interest in the computational modeling of semantic change. Thus far, the research community has primarily focused on identifying words that have changed their meaning over time. Existing approaches are primarily based on vector token representations (embeddings) and thus often do not enable the \textit{interpretation} of the novel senses a word has gained \cite{10.1145/3672393}. Only recently, with the advent of new generative language models, the research community has begun to turn its attention to the interpretation of detected semantic change. One step in this direction was the AXOLOTL’24 shared task on explainable semantic change modeling \cite{fedorova-etal-2024-axolotl24}.

The shared task was focused on the analysis of diachronic semantic shifts between two time periods, a challenge typical for historical linguists and lexicographers. It consisted of two separate subtasks, given a set of target word $x$ usages (examples):

\begin{enumerate}[nosep]
    \item find the usages of $x$ in novel senses;
    \item provide human-readable descriptions (such as definitions) of the novel senses.
\end{enumerate}

In this paper, we apply LLM-based definition generators \cite{noraset2017definition,gardner2022definition,segonne-mickus-2023-definition} to the second subtask of AXOLOTL'24, where the participants were asked to create  descriptions or definitions for novel word senses.  `Novel' here means a sense which is present in a corpus from the newer (`second') time period,  but is not mentioned in a dictionary covering the older (`first') time period. In the simplest form, the task is to provide a correct definition of a novel sense $y$ of a target word $x$, given a bunch of $x$ usages belonging to $y$. The performance of the systems was evaluated with BLEU \cite{papineni-etal-2002-bleu} and BERTScore \cite{Zhang2020BERTScore}, comparing the generated definitions to manually annotated gold definitions. An example instance from the AXOLOTL'24 shared task is given in Table~\ref{tab:example} in the Appendix. 

The AXOLOTL'24 shared task was offered in three languages: Finnish, Russian and German, with the latter as a `surprise language' without training and validation datasets. Three teams participated in subtask 2, achieving promising performance, but still leaving ample room for improvements. One of the teams used  GlossBERT \cite{huang-etal-2019-glossbert} fine-tuned with adapters to match usage examples to senses and definitions retrieved from Wiktionary. Two other teams prompted the GPT 3.5 language model to generate senses and definitions. 

Thus, the best-performing systems relied on a closed proprietary LLM and a lexical database, respectively. However, the use of a closed model is not ideal as it lacks transparency, and limits accessibility for researchers. Similarly, the use of existing lexical resources such as WordNet or Wiktionary at test time contradicts the goal of identifying novel senses, as genuinely new meanings are, by definition, absent from established ontologies. For this reason, in this paper we concentrate on the generative approach only and evaluate definition generators built on three different instruction-tuned open-weights LLMs: \texttt{mT0}, \texttt{Aya-101} and \texttt{TowerInstruct}.
Our contributions are as follows:

\begin{enumerate}
\item We provide open alternatives to the generative systems used by the AXOLOTL'24 participants.
\item We explore the differences between performance of models sharing the same architecture, pretraining procedure (base pretraining with further instruction tuning) and base pretraining data, but having different number of parameters and sizes of the instruction datasets (\texttt{mT0} and \texttt{Aya-101}).
\item We investigate how the outputs from fine-tuned encoder-decoder (\texttt{mT0} and \texttt{Aya-101}) and decoder-only (\texttt{TowerInstruct}) models differ both in terms of automatic metrics and human evaluation. 
\item We investigate how much fine-tuning data is needed for a reliable definition generator.
\end{enumerate}

Our code is publicly released on GitHub\footnote{\url{https://github.com/ltgoslo/MultilingualDefGen}} and model adapters are available on HuggingFace\footnote{\url{https://huggingface.co/collections/ltg/definition-modeling-6580c4598ecea67c7d5b1970}}.

\section{Data}
\label{sec:data}

In order to adapt an existing generative LLM for the task of generating definitions, one needs to fine-tune it on a corresponding dataset. In this work, we utilize two resources: the AXOLOTL'24 training data, and definitions and examples from Dbnary \cite{serasset-2012-dbnary}, a lexicographic resource derived from Wiktionary.

For Finnish and Russian, we fine-tune two versions of each model:

\begin{enumerate}[nosep]
    \item on AXOLOTL'24 data only \textbf{(a)},
    \item on a combination of AXOLOTL'24 and Dbnary data \textbf{(a+d)}.
\end{enumerate}

\noindent Since no German training data was provided in AXOLOTL'24, we use only Dbnary for fine-tuning the German models.

Some of the AXOLOTL'24 test set senses and definitions ultimately come from Wiktionary. Thus, to avoid data contamination, 
we removed from our Dbnary datasets \textit{all the words (along with their senses, definitions and usages) also present in the AXOLOTL'24 test and development sets for all languages}.

Table~\ref{tab:data} shows the sizes of data splits in our experiments (in example-definition pairs).  
The validation set for German comes from Dbnary, while the validation sets for Russian and Finnish come from AXOLOTL'24. Test sets for all languages come solely from AXOLOTL'24 \citep[for more statistics, see Section 3 and Appendix A3 of][]{fedorova-etal-2024-axolotl24}.

\begin{table}[ht]
\centering
\begin{tabular}{r|ccc}
\toprule
\textbf{Split} & \textbf{Russian} & \textbf{Finnish} & \textbf{German} \\
\midrule
Train \textbf{(a)} & 6,494 & 93,139 & -- \\
Train \textbf{(a+d)} & 180,072 & 119,980 & 322,937 \\
Dev \textbf{(a)} & 2,026 & 6,554 & -- \\
Dev \textbf{(d)} & -- & -- & 19,398 \\
Test \textbf{(a)} & 2,126 & 6,725 & 1,152 \\ 
\bottomrule
\end{tabular}
\caption{Data split sizes used in our experiments (example-definition pairs). \textbf{a} stands for AXOLOTL'24, \textbf{d} stands for Dbnary.}
\label{tab:data}
\end{table}

Dbnary greatly increases the amount of fine-tuning data. 
As shown in Section~\ref{sec:subtask2}, including it  
improves the performance.

\section{Definition generators}

This section motivates the choice of models to fine-tune and describes the training setup.

\subsection{TowerInstruct}
\texttt{TowerInstruct-7B}\footnote{\url{https://hf.co/Unbabel/TowerInstruct-7B-v0.2}} is Llama-2-7B decoder-only model \cite{touvron2023llama2openfoundation} enhanced with continued pretraining on a mix of 20 billion tokens of monolingual data in ten different languages -- English, Portuguese, Spanish, French, German, Dutch, Italian, Korean, Chinese, and Russian -- as well as other bilingual data (including Finnish) \cite{alves2024tower} and further fine-tuned on instructions relevant for translation. The choice of a Llama-based model is motivated by its usage in previous works \citep{periti-etal-2024-automatically}. We refer to our models fine-tuned from it as \texttt{TowerDictionary}.

\subsection{mT0-XL}

\texttt{mT0-XL}\footnote{\url{https://hf.co/bigscience/mT0-xl}} is a version of the multilingual encoder-decoder \texttt{mT5} model (originally pre-trained on 108 languages) that was instruction fine-tuned on the \texttt{xP3} dataset, containing instructions for 13 training tasks in 46 languages with English prompts \cite{muennighoff-etal-2023-crosslingual}.
This model is 3.7B parameters in size, about half the size of \texttt{TowerInstruct-7B}. This model was also used in previous works on fine-tuning definition generators \cite{giulianelli-etal-2023-interpretable, kutuzov-etal-2024-enriching, fedorova-etal-2024-definition}.
We refer to our models fine-tuned from it as \texttt{mT0DefGen}.

\subsection{Aya-101}

\texttt{Aya-101}\footnote{\url{https://hf.co/CohereForAI/aya-101}} has the same architecture and pretraining dataset as \texttt{mt0-XL} (and even reuses its tokenizer), but is larger in size (13B parameters) and instruction-tuned on 101 languages, including our three languages of interest \cite{ustun-etal-2024-aya}.\footnote{We do not use its successor \texttt{Aya-23}, since it was not optimized for Finnish.} We refer to our fine-tuned models as \texttt{Aya-101-DefGen}.

\subsection{Instruction fine-tuning}

We employed Quantized Low-Rank Adaptation (QLoRA) \cite{dettmers2023qlora} applied to all linear layers of the models. Hyperparameters are provided in Appendix~\ref{sec:parameters}. For each language, the same prompt (to be found in Appendix~\ref{sec:prompts}) was used in all experiments. We compared several generation strategies and chose beam search for all models.

\subsection{Aggregating different definitions for the same sense}

The fine-tuned models generate one definition per usage example, but this is not sufficient to solve AXOLOTL'24 subtask 2, which expects predictions in the form of `target--sense--definition' triplets. Since the AXOLOTL'24 datasets contain more than one usage per sense in most cases, this means that the definitions generated for all usages of one sense must be `aggregated' to produce a single definition (the `sense label'). 

We implemented this aggregation in a very straightforward manner inspired by  \citet{giulianelli-etal-2023-interpretable}. Definitions for all usages of a given sense are embedded using the Sentence Transformers model\footnote{\url{https://hf.co/sentence-transformers/distiluse-base-multilingual-cased-v1}} \cite{reimers-gurevych-2020-making}. Then, the `prototypical embedding' is found by computing the average of all definition embeddings, and the definition with the embedding closest to the prototypical one (by cosine similarity) is chosen as the sense label.

In addition, we ensure that the definitions are unique across senses: we try to avoid cases when two different senses $a$ and $b$ are assigned one and the same sense label. For this purpose, before assigning a sense label we check whether this definition has already been assigned to another sense of the same target word. If the answer is positive, the current sense label candidate is discarded, and the definition next closest to the prototypical embeddings is chosen. Thus, we loop over candidate definitions sorted by their frequency until a generated definition is found which has not been assigned to any sense yet. Only if no such definition is found among the usages of the current sense, we fall back to the most prototypical definition (this results in non-unique sense labels). In our experiments, using this technique resulted in small but consistent improvements across all models and languages.

We have done preliminary experiments to see if instruction-tuned models could summarize the definitions of the same sense, but they turned out to be insufficient to solve this task.

\section{Results}
\label{sec:subtask2}

The fine-tuned definition generators were used to create definitions for the usages with novel senses from the AXOLOTL'24 subtask 2 test sets, which were then aggregated as described above. In this setup, we assumed that subtask 1 (finding these usages) is already solved. Quoting the shared task organizers, ``\textit{the evaluation of Subtask 2 therefore limits itself to evaluating the validity of provided glosses}" \cite{fedorova-etal-2024-axolotl24}.

The resulting `target--sense--definition' triplets were evaluated by the official AXOLOTL'24 scoring code.\footnote{\url{https://github.com/ltgoslo/axolotl24_shared_task/blob/main/code/evaluation/scorer_track2.py}} Table~\ref{tab:task2_2} reports the performance of \texttt{TowerDictionary} \texttt{mT0DefGen} and \texttt{Aya-101-DefGen} models, as well as the best results achieved by \textit{generative} models for each language from the AXOLOTL'24 leaderboard. We report BLEU and BERTScore, both in the range of $0-100$.

\begin{table*}
    \centering
\resizebox{0.95\linewidth}{!}{
    \begin{tabular}{l|lll}
    \toprule
     \textbf{Model} & \textbf{Russian}  & \textbf{Finnish}  & \textbf{German} \\
         \midrule
    TowerDictionary \textbf{(a)}   & 3.98 / 65.64 / 64.98 / 65.22   & \textbf{5.53} / 66.19 / 67.58 / \textbf{66.81}  & -- / -- \\
    TowerDictionary \textbf{(a+d}) & \textbf{7.22} / 69.49 / 68.78 / \textbf{69.04}  & 4.57 / 65.57 / 67.34 / 66.37  & \textbf{2.25} / 64.26 / 68.27 / \textbf{66.12} \\
    \midrule
    mT0DefGen \textbf{(a)}    & 6.46 / 66.97 / 65.84 / 66.31  & 4.54 / 63.98 / 66.54 / 65.17  &  -- / --\\
    mT0DefGen \textbf{(a+d)}  & 6.01 / 67.95 / 66.25 / 67.00  & \textbf{5.54} / 64.72 / 66.67 / 65.61 & \textbf{2.28} / 63.07 / 65.87 / \textbf{64.38}  \\
    \midrule
    Aya-101-DefGen \textbf{(a)} &  6.69 / 68.50 / 67.74 / 68.00  &  \textbf{5.99} / 66.85 / 69.00 / \textbf{67.83}   & -- / -- \\
    Aya-101-DefGen \textbf{(a+d)} & \textbf{6.37} / 68.98 / 67.57 / \textbf{68.18}  & 5.15 / 66.05 / 68.08 / 66.98 &  \textbf{2.39} / 64.33 / 67.74 / \textbf{65.96} \\
    \midrule
    Best AXOLOTL'24 & 2.68 / -- / -- / 65.64 & 2.32 /  -- / -- / 67.46 & 1.00 /  -- / -- / 65.24 \\
    \bottomrule
    \end{tabular}
    }
    \caption{AXOLOTL'24 subtask 2 scores (BLEU / BERTScore precision * 100 / BERTScore recall * 100 / BERTScore F1 * 100); `\textbf{a}' stands for `fine-tuned on AXOLOTL', `\textbf{a+d}' for `fine-tuned on AXOLOTL+Dbnary'. Best AXOLOTL'24 are the best \textit{generative} approaches used by participants. The best results are highlighted with bold (may be more than one for a language, if the difference is not statistically significant.)}
    \label{tab:task2_2}
\end{table*}

\paragraph{The winner} Definition generators fine-tuned on open-weights LLMs outperform the best AXOLOTL'24 Subtask 2 \textit{generative} submissions for all three languages under analysis. While \texttt{TowerDictionary} and \texttt{Aya-101-DefGen} perform better than \texttt{mT0DefGen}, it is not possible to define a winner based only on BLEU and BERTScore, since differences between \texttt{TowerDictionary} and \texttt{Aya-101-DefGen} are not statistically significant (as per t-test\footnote{\url{https://docs.scipy.org/doc/scipy/reference/generated/scipy.stats.ttest_ind.html}}). For this reason, we have conducted a manual error analysis.

\subsection{Qualitative analysis}

\begin{table*}
    \centering
\resizebox{0.9\linewidth}{!}{%

    \begin{tabular}{l|rrrrrrrrr}
    \toprule
        \textbf{Model} & \multicolumn{3}{c}{\textbf{Russian}} &  \multicolumn{3}{c}{\textbf{Finnish}} & \multicolumn{3}{c}{\textbf{German}} \\
        & Fluency & Adeq. & Circ. & Fluency & Adeq. & Circ. & Fluency & Adeq. & Circ. \\
        \midrule
       TowerDictionary (a)  & \textbf{18.8} & 40.6 & 15.5 & 13.33 & 86.67 & 15.1 & -- & -- & --  \\
       TowerDictionary (a+d)  & 53.1 & 50.0 & 18.4 & \textbf{6.67} & 76.67 & \textbf{14.8} &  \textbf{3.9} & 57.7 & \textbf{3.3} \\
       \midrule
       mT0DefGen (a)  & 87.5 & 50.0 & 15.6 & 43.33 & 0.8  & 21.5 & -- & -- & -- \\
       mT0DefGen (a+d)  & 56.3 & 43.8 & 26.8 & 66.67 & 86.67 & 21.5 & 30.8 & 76.9 & 4.9 \\
       \midrule
       Aya-101-DefGen (a)  & 90.6 & 40.6 & \textbf{10.6} &  26.67 & 83.33 & 19.2 & -- & -- & -- \\
       Aya-101-DefGen (a+d)  & 43.8 & \textbf{37.5} & 19.9 & 33.33 & \textbf{73.33} & 20.7 & 15.4 & \textbf{53.9} & 4.9 \\
       \bottomrule
    \end{tabular}
    }
    \caption{Share of definitions with fluency-related issues, adequacy-related issues, and containing circularity (\%) -- lower values are better. `a' stands for `fine-tuned on AXOLOTL', `a+d' for `fine-tuned on AXOLOTL+Dbnary'.}
    \label{tab:issues}
\end{table*}

We annotated the generated definitions according to three criteria:

\begin{enumerate}
\item The definition has \textbf{fluency issues} \cite{snover-etal-2006-study}: it contains repetitions, wrong dictionary labels such as \textit{`colloquial'} or \textit{`metaphoric'}, wrong punctuation, or the sentence is grammatically incorrect.
\item The definition has \textbf{adequacy issues}: the definition contains factual mistakes (e.g., \textit{`Eurasian jay: a bird of the Felidae family'}), it refers to the incorrect sense of the word, or is too broad (e.g., \textit{`Eurasian jay: a small bird'}) or too narrow.
\item The definition is \textbf{circular}: the generated definition contains the target word (`definiendum') itself (e.g., \textit{`a table is a sort of a table'}).
\end{enumerate}

Fluency and adequacy issues were annotated manually on random samples of definitions for Finnish (30 samples) and Russian (32 samples), and on the entire test set for German (26 samples). Circularity was detected automatically on the full test sets. Table \ref{tab:issues} shows the results of the error analysis.

\paragraph{3.7B vs 13B parameters} 

The comparison of \texttt{mT0DefGen} and \texttt{Aya-101-DefGen} shows that \texttt{mT0DefGen} more often generates semantically incorrect definitions  and is more prone to repetitions. A German example: for the sense \textit{`verringern, reduzieren'} (cut down, reduce) of the target word \textit{`abbauen'} \texttt{mT0DefGen} generated \textit{`Transitiv: etwas entfernen, entfernen'} (transitive: to remove something, remove), while \texttt{Aya-101-DefGen} generated \textit{`Transitiv: etwas Transitiv: etwas reduzieren, verringern'} (transitive: to reduce something, to cut down). For the sense \textit{`etw. Aufgebautes (z.B. Kr\"amerbude) zerlegen, abbrechen'} (to disassemble, to demolish smth. built (e.g. grocer's shop)) \texttt{mT0DefGen} generated \textit{`Transitiv, auch reflexiv:; etwas reduzieren, verringern'} (transitive, also reflexive:;to reduce something, to cut down), while \texttt{Aya-101-DefGen} output \textit{`Transitiv: etwas entfernen, zerstören'} (transitive: to remove something, to destroy). Thus, outputs of \texttt{Aya-101-DefGen} are adequate, while senses in \texttt{mT0DefGen}'s outputs are swapped. Therefore, we recommend to prefer \texttt{Aya-101-DefGen} upon \texttt{mT0DefGen} despite its larger size.

\paragraph{Encoder-decoder vs. decoder-only} 

As for the model architecture itself (encoder-decoder or decoder-only), our results correspond with the findings of the related works that both types of models are suitable for the task. However, \texttt{TowerDictionary} performs better in terms of fluency, while \texttt{Aya-101-DefGen} provides better adequacy. There is no clear winner in terms of circularity.

\paragraph{Amount of fine-tuning data} Augmenting the AXOLOTL'24 training set with Dbnary always improves the results of Russian models in BERTScore. While BLEU on AXOLOTL-only data is higher for encoder-decoder models, the manual inspection shows that Russian models trained on AXOLOTL-only data overfitted to excessively output dictionary labels, which explains their low fluency. Since these labels are common in gold data, they boost BLEU, but do not add much to understanding the word's semantics and may cause higher metrics even if the sense is wrong. Thus, for the Russian split, BERTScore is a more reliable metric and the dataset of 7K instances was not sufficient.

For Finnish, the results are controversial across the models. The reason might be very different sources of data: while in case of Russian both AXOLOTL and Dbnary are sampled from Wiktionary, in case of Finnish AXOLOTL data were borrowed from a historical dictionary. Thus, the data domain should be still preferred over data quantity when training definition generators. This observation also holds for German: its ground truth definitions are also not from Wiktionary and completely lack dictionary labels, while the output of both \texttt{mT0DefGen} and \texttt{Aya-101-DefGen} is full of `metaphoric' etc. 

\paragraph{Circular definitions}
For Russian, \texttt{Aya-101-Def-} \texttt{Gen} avoids circularity better than the other two models. For German and Finnish, encoder-decoder models are less prone to circularity than the decoder-only model. Also, fine-tuning on Dbnary has different effects across languages, again proving the importance of high-quality training data.

\subsection{Comparison with AXOLOTL'24 submissions}

We compared definitions generated by our models to those of the AXOLOTL'24 Wooper-NLP team, which submitted predictions for the highest number of target words. The first notable difference is that Wooper-NLP's definitions are twice as long as the gold answers in terms of character counts, while the length of the generations from our models is better aligned with the ground truth. We also looked through the examples described as problematic in the shared task paper \citep[][Appendix C6]{fedorova-etal-2024-axolotl24}. The definitions generated by our fine-tuned models seem to avoid the problem that `\textit{the model doesn’t stop after producing the definition, but continues with an explanation or excessive details}'. Also, our definitions are not overly narrow because of repeating named entities from the usage examples (this explains why GPT3.5’s definitions are too long and ours are not).  The predictions of the fine-tuned models are also  less prone to grammatical and spelling errors, and loan translations, which proves that large proprietary models may still have issues with language-specific generation; fine-tuning of open models makes sense even for large and mid-resource languages.

\section{Conclusion}

In this work, we use large language models fine-tuned on definition modeling to generate labels for novel senses. One can think of this task as updating a dictionary based on a set of new texts.

The participants of the AXOLOTL'24 shared task mostly tackled the problem by using the proprietary GPT 3.5 model. We show that one can instead fine-tune \textit{open-weights} LLMs and achieve a better performance in this subtask with contextualized definitions generated by them. We also publicly share the models.

Our comparison of different base models showed that instruction-tuned encoder-decoder (T5-like) models perform on par with their decoder-only (Llama-like) counterparts. According to automatic metrics, larger models always outperform smaller ones, as well as larger fine-tuning datasets often do. However, not only data quantity, but also its relevance to the domain matters. Our experiments also show that few thousands of fine-tuning data instances may not be enough for a complex semantic task such as definition generation.

\section*{Limitations}
An obvious limitation of this paper is its focus on the second subtask of AXOLOTL'24 only. This means we are dealing with `given' novel senses, without the need to actually identify them, and the evaluation is focused on the ability of the systems to produce a sensible sense definition from a set of usages. We leave the challenge of developing an end-to-end approach that solves both subtasks jointly for future work.

Language-specific hyperparameter choice and data preprocessing such as dealing with too long texts and removing special dictionary labels is beyond the scope of this paper. Instead, we try to make language-independent observations.

It is likely that all three base models have been exposed to the Russian test set, which is taken from Wiktionary. However, they produce  
predictions that are very different from the ground truth, which is not surprising, since definition modeling task was not among their instructions.

It is also important to note that fine-tuned definition generators inherit the biases and peculiarities of the lexical resources they were trained on, which can become a potential risk.

\bibliography{anthology,custom}

\appendix

\section{Instruction tuning hyperparameters}
\label{sec:parameters}

All models were fine-tuned for 1 epoch with the settings and hyper-parameters shown in Table~\ref{tab:params}.

\begin{table}[!ht]
\begin{tabular}{lccc}
\toprule
\multicolumn{1}{l|}{\textit{Weight decay}} & \multicolumn{3}{c}{Tower 0.001, mt0 and Aya 0} \\ 
\multicolumn{1}{l|}{\textit{Learning rate}} & \multicolumn{3}{c}{Tower 1e-4, mt0 and Aya 5e-5} \\ 
\multicolumn{1}{l|}{\textit{Warmup ratio}} & \multicolumn{3}{c}{0.05} \\ 
\multicolumn{1}{l|}{\textit{Batch size}} & \multicolumn{3}{c}{16} \\ 
\multicolumn{1}{l|}{\textit{Optimizer}} & \multicolumn{3}{c}{paged\_adamw\_8bit} \\ 
\multicolumn{1}{l|}{\textit{LoRA rank}} & \multicolumn{3}{c}{256} \\ 
\multicolumn{1}{l|}{\textit{LoRA alpha}} & \multicolumn{3}{c}{512} \\ 
\multicolumn{1}{l|}{\textit{LoRA dropout}} & \multicolumn{3}{c}{0.1} \\ 
\bottomrule
\end{tabular}\
\caption{Settings and hyper-parameters for model fine-tuning.}
\label{tab:params}
\end{table}

We compared performance of models trained with \texttt{adafactor}\footnote{\url{https://huggingface.co/docs/transformers/main/en/main\_classes/optimizer\_schedules\#transformers.Adafactor}} and \texttt{paged\_adamw\_8bit}\footnote{\url{https://huggingface.co/docs/bitsandbytes/reference/optim/adamw\#bitsandbytes.optim.PagedAdamW8bit}} optimizers on the development set and found no significant difference in metrics according to t-test, while the latter required less runtime.

All models were trained on a single AMD MI250x (64 GB) GPU, except for Finnish TowerDictionary models, which were trained on a single NVIDIA A100-SXM4-80GB. The training time varied from hours to two days for the largest combination of model parameters and training data, Aya-101 fine-tuned on German Dbnary. 

Using weight decay allowed to pass the whole data, while without it the training early stopped after several hundred steps.

\section{Generation hyperparameters}

Table \ref{tab:gen-params} specifies generation parameters used to obtain results presented in the table \ref{tab:task2_2}.

\begin{table}[!h]
\resizebox{1\columnwidth}{!}{
\begin{tabular}{ll}
\toprule
\textit{num\_beams} & 5 \\ 
\textit{do\_sample} & False \\ 
\textit{length\_penalty} & 1.1 \\ 
\textit{early\_stopping} & True \\ 
\textit{repetition\_penalty} & 
1.1 \\ 
\bottomrule
\end{tabular}\
}
\caption{Settings and hyper-parameters for text generation.}
\label{tab:gen-params}
\end{table}

We compared different generation strategies\footnote{\url{https://huggingface.co/docs/transformers/en/main_classes/text\_generation\#transformers.GenerationConfig}} both for an encoder-decoder and a decoder-only model on the Russian development split.
The exact parameters of various generation strategies that we tried are available in our Github repository\footnote{\url{https://github.com/ltgoslo/MultilingualDefGen/blob/dff165051166b3bdf2a6dedd07904c99868f47ad/src/modeling/decoder\_only\_predict.py\#L25}}. The results can be found in Table \ref{tab:gen-params-dev}. The models are ``sensitive'' to generation parameters. The best results obtained from beam search combination with multinomial sampling and contrastive search\footnote{\url{https://huggingface.co/blog/introducing-csearch}} may be explained by the known fact \cite{welleck2019neural} that non-deterministic decoding ``suffers'' less from repetitions. However, to ensure reproducibility of our results, we have chosen the standard beam search for our experiments (also the same paper argues that ``being prone'' to repetitions may still depend more on how a model was trained rather than on decoding strategies).

\begin{table}[!h]
\resizebox{1\columnwidth}{!}{
\begin{tabular}{lll}
\toprule
Strategy & mT0 & Tower \\
\textit{greedy search} & - & 63.52 \\ 
\textit{multinomial sampling} & - & 62.8 \\ 
\textit{beam search} & - & 64.15\\ 
\textit{beam search multinomial sampling} & \textbf{67.11} & \textbf{64.39} \\ 
\textit{contrastive search, repetition penalty 1.1} & 66.87 & 64.33 \\ 
\textit{contrastive search, repetition penalty 1.2} & 61.66 & 64.05 \\
\textit{dola decoding} & - & 63.83 \\ 
\bottomrule
\end{tabular}\
}
\caption{Different decoding strategies, Russian development set.}
\label{tab:gen-params-dev}
\end{table}

\section{Prompts}
\label{sec:prompts}

Following \citet{giulianelli-etal-2023-interpretable}, the model input was formatted like a usage example followed by a prompt that can be roughly translated to English as `What is the meaning of <target word>?'. The prompts were suggested by human speakers of the corresponding languages and are reported in the Table~\ref{tab:mt0-prompts}. These prompts were used both for fine-tuning and for inference.

\begin{table}[!ht]
\resizebox{1\columnwidth}{!}{
\begin{tabular}{ll}
\toprule
\textbf{Language} & \textbf{Prompt} \\
\midrule
Russian & \foreignlanguage{russian}{Что такое <target word>?} \\
Finnish & . Mitä tarkoittaa <target word>? \\
German & . Was ist die Definition von <target word>? \\
\bottomrule
\end{tabular}
}
\caption{Prompts for the definition generation models. We also experimented with `Mikä on <target word>?' for Finnish, but it caused model to generate noun-like definitions for the target words in other parts of speech. This caused lower scores, so we do not report it in the main text.}
\label{tab:mt0-prompts}
\end{table}

\section{Example of an AXOLOTL'24 instance}

\begin{table}[h]
\begin{tabular}{lp{52mm}}
\toprule
\textbf{Target}: & cell \\
\textbf{Sense}: & \textsc{cell\_3} \\
Period: & new \\
Usage: & In multicellular organisms, groups of \textit{cells} form tissues and tissues come together to form organs \\
\textbf{Definition}: & A unit of a living organism \\
\bottomrule
\end{tabular}
\caption{An example instance of the AXOLOTL'24 training set in English. In the test set, the \textit{definition} field for the usages from the `new' time period is blank. For subtask 2, the participants have to submit \textit{(target, sense, definition)} triplets for each novel sense of the target words.}
\label{tab:example}
\end{table}

For subtask 2, the test submission must consist of `target-sense-definition' triplets (bold in the table~\ref{tab:example}).

\section{Peculiarities of the German dataset}

The lower metrics in Table \ref{tab:task2_2} for German might be explained by a large share of test instances consisting of many sentences, while all data splits for other languages and German Dbnary mostly feature usage examples not longer than one sentence. In order to avoid truncating usage examples before the target word occurrence, we splitted the text into sentences and selected only those containing the target word lemmas. We used \texttt{SpaCy}\footnote{\url{https://spacy.io/usage/models}} model \texttt{de\_core\_news\_sm} for that.

\end{document}